\documentclass{article} 
\usepackage{iclr2027_conference,times}

\usepackage{amsmath,amsfonts,bm}

\def\eqref#1{equation~\ref{#1}}

\def\1{\bm{1}}

\DeclareMathAlphabet{\mathsfit}{\encodingdefault}{\sfdefault}{m}{sl}
\SetMathAlphabet{\mathsfit}{bold}{\encodingdefault}{\sfdefault}{bx}{n}

\usepackage{hyperref}
\usepackage{url}
\usepackage{graphicx}
\usepackage{booktabs} 
\usepackage[table]{xcolor}
\usepackage{colortbl}
\usepackage{microtype}

\title{Dataset Biases and Shortcut Learning in Motion-Based AI-Generated Video Detection}

\author{Joren Michels, Lode Jorissen \& Nick Michiels \\
Digital Future Lab - Flanders Make - Hasselt University\\
}

\iclrfinalcopy 
\begin{document}

\maketitle

\begin{abstract}
The visual quality of AI-generated videos has improved drastically in recent years, making it increasingly difficult for humans to distinguish between real and synthetic media. In this work, we evaluate the robustness and applicability of four state-of-the-art motion-based AI-generated video detectors. We identify significant preprocessing and sampling biases in three of the four methods and demonstrate that they account for a substantial portion of their reported performance. Furthermore, we find that these detectors are highly sensitive to motion patterns specific to their evaluation datasets, where AI-generated videos generally exhibit less inter-frame movement than real videos. We show that for all detectors, performance collapses to near-random levels when evaluated on a dataset that does not contain this motion bias. Additionally, through dataset rebalancing and the application of simple spatial augmentations, we observe severe performance degradation across all evaluated models. In contrast, we find that an existing frequency-based detector maintains strong performance across all evaluated datasets, suggesting that frequency-based approaches may offer a more generalizable path forward for AI-generated video detection. We hope that our work raises awareness towards these vulnerabilities and encourages the development of more representative, unbiased datasets and more robust evaluation protocols.
\end{abstract}

\section{Introduction}
\label{sec:introduction}

The rapid advancement of generative video models has resulted in AI-generated videos that are increasingly more realistic \citep{generation1, generation2, generation3, Veo3, Sora2}. It has already been shown that humans  cannot differentiate reliably anymore between real and synthetic media \citep{ai-video-human-detection}. Although there are multiple interesting use cases for AI-generated media, for example in the creative sector, the dangers of this technology cannot be underestimated. One possible danger is the generation of media content that can be perceived as being harmful or insulting \citep{unsafe-difusion}. AI-generated media can also be used to mislead the public, fraud people, and sway public opinion \citep{ai-typology-risks}. The fact that exposure to AI-generated content can even distort human memory, amplifies these societal dangers even more \citep{ai-video-memory}. These risks necessitate the development of reliable AI-generated video detectors that can be deployed in real-world scenarios.

Research on this topic was originally focused on the detection of deepfakes (AI-generated videos of human faces)\citep{deepfake1, deepfake2}. These methods often focus on subtle facial inconsistencies, limiting their applicability outside of this specific domain. Only recently, with the upcoming of diffusion-based generative video models, has the direction also shifted towards the detection of more general AI-generated videos. These detectors rely on a range of techniques that can broadly be divided into three main categories, with some methods combining multiple strategies. A common approach involves training a machine learning model on a large dataset of videos \citep{demamba, AIGVDET, waverep, NSGVD, UNITE, Text-Vision, TurnsOutImNotReal}. Such methods often enhance their inputs using derived representations of the original videos, such as optical flow maps \citep{AIGVDET}, reconstruction errors obtained from pre-trained diffusion models \citep{TurnsOutImNotReal}, or modifications to specific frequency bands \citep{waverep}. A second category of detectors integrates large language models (LLMs) into their frameworks, using them not only for detection but also to enhance the explainability of the predictions \citep{MMDet, Vidguard, buster}. Lastly, a third set of techniques look at inter-frame differences to detect AI-generated videos based on unnatural motion. These approaches either train models to recognize such motion irregularities \citep{restrav, NSGVD} or detect them using training-free methods \citep{D3, Over-Coherence}. Inter-frame differences are mostly calculated in the feature space of a pre-trained feature encoder, like DINOv2 \citep{DINOv2} or XCLIP \citep{xclip}.

We focus on this third category of motion-based detectors. These methods are recent and are already being taken up in subsequent work, yet their robustness has not been systematically examined, which motivates the analysis presented here. In particular, we evaluate four recent methods: \emph{D3} \citep{D3}, \emph{ReStraV} \citep{restrav}, \emph{Over-Coherence} \citep{Over-Coherence} and \emph{NSG-VD} \citep{NSGVD}. Reported results for three of these approaches indicate near-perfect performance on their evaluation datasets (D3: 97.72 AUC, ReStraV: 98.81 mAP, NSG-VD: 96.14 AUC). However, we identify significant sampling and preprocessing biases underlying these three methods, which they might inadvertently exploit. We systematically analyze the extent of these biases and evaluate how detection performance changes once they are removed. Our findings indicate that a substantial portion of the reported performance of these detectors stems from exploiting such biases, rather than capturing motion-related differences between AI-generated and real videos. 

Furthermore, our analysis reveals that the high accuracy of these detectors also depends on motion biases embedded in recent evaluation datasets. An analysis of five recent datasets \citep{demamba, buster, waverep, Over-Coherence, vidprom} suggests a noticeable imbalance in motion patterns: in four of the five datasets, AI-generated videos tend to exhibit limited inter-frame movement, while real videos more often contain dynamic motion and scene transitions. As generative models rapidly evolve, such biased datasets fail to accurately represent the current capabilities of AI video generation. When we re-evaluate the detectors on temporally (re-)balanced datasets, their performance degrades significantly. We also demonstrate that this over-reliance on motion biases exposes a critical vulnerability. By applying targeted spatial augmentations that cause minimal perceptual degradation, accuracy is drastically reduced. To test if these problems are specific to motion-based detectors, we also evaluate a non-motion-based detector (\emph{WaveRep} \citep{waverep}) on the same five datasets used in our main evaluation. In contrast to the motion-based models evaluated, WaveRep maintains high accuracy across all datasets, suggesting that frequency-based detectors do not rely on the motion bias that affects motion-based detectors. Finally, based on the learned insights, we discuss practical implications for the construction of evaluation datasets and the design of more reliable detection benchmarks. \textbf{To conclude, this work is an audit rather than a new method: we stress-test the evaluation pipelines behind recently reported results and show how much of their performance reflects dataset and preprocessing biases rather than genuine detection capabilities.}

\section{Related Work}
\label{sec:related_work}

\paragraph{AI video detectors}
Detecting AI-generated videos has become a crucial task given the rapid advancement of generative video models. \citet{AIGVDET} train a spatial domain detector and an optical flow detector and combine their outputs on a per-frame basis. \citet{waverep} propose a data augmentation strategy by replacing specific frequency bands of frames of fake videos with real ones using wavelet decomposition, after which a DINOv2 model \citep{DINOv2} is trained on this augmented dataset. Other approaches train spatio-temporal models. For instance, \citet{TurnsOutImNotReal} train a CNN+LSTM architecture on sequences of video frames. \citet{Text-Vision} leverage a pre-trained video feature encoder and train a transformer model on the extracted features. Similarly, \citet{UNITE} use a pre-trained image feature encoder together with a transformer architecture and introduce an attention-diversity loss. \citet{demamba} train a Mamba-style architecture on a large-scale dataset of real and AI-generated videos. A second set of detectors use Multimodal Large Language Models (MLLM's) as (part of) their method to detect AI-generated videos. \citet{MMDet} train a dynamic fusion head to combine feature outputs of an LLM branch and a trainable spatio-temporal branch, while \citet{Vidguard} and \citet{buster} use reinforcement learning to improve reasoning capabilities. Lastly, multiple recent papers leverage differences between feature embeddings of subsequent frames (or direct frame differences) as a cue to detect AI-generated videos. \citet{D3} take the standard deviation of the second order differences of feature embeddings of subsequent frames as a metric for detection. \citet{Over-Coherence} follow a similar approach by computing the cosine similarity between subsequent frame embeddings. Based on the perceptual straightening hypothesis, \citet{restrav} train an MLP on the distances and angles between embeddings of subsequent frames. Lastly, \citet{NSGVD} combine the results of the score function of a pre-trained diffusion model and inter-frame differences in a single metric that is compared against a reference set of real videos. These four methods are the ones that will be evaluated in this paper.

\paragraph{Biases and shortcut learning}

Research about dataset biases and possible shortcut learning as a result has been conducted way before the upcoming of modern generative video models. Early work on images has repeatedly shown that models can achieve strong performance by exploiting dataset biases rather than learning the intended task. For example, \citet{pitfall} showed that a widely used dataset for tampered image detection contains a bias related to JPEG compression: untampered images were saved at different JPEG quality factors than tampered images. They observed significant performance drops when these biases were removed. Similarly, \citet{kinship} uncovered a bias in a kinship verification dataset, where face crops of related individuals were consistently extracted from the same original image. As a result, paired images shared low-level characteristics unrelated to kinship, enabling even a simple chrominance distance metric to rival state-of-the-art methods in detecting if two people are related. Later, \citet{criminality} questioned the high accuracies of an algorithm that detects if someone is a criminal based on an image of their face. They identified multiple biases in the evaluation dataset, including differences in camera model and image format (JPEG vs PNG) between the two classes. More recently, \citet{cameraidentification} showed that a classifier trained for camera identification relied heavily on scene content and color distributions, since images from the same camera were often captured under similar conditions, rather than true sensor-specific fingerprints. These examples highlight a long known issue in machine learning: models tend to look for the easiest way to complete a task. If not enough care is taken in constructing unbiased datasets, a model might exploit biases not related to the task at hand, also known as shortcut learning, leading to misleadingly high performance that does not generalize beyond the biased data.

Unfortunately, these types of biases are also present in the field of AI-generated media detection. Several studies have identified differences in JPEG compression within commonly used detection datasets: real images are frequently JPEG-encoded, whereas AI-generated images are not \citep{fakeorjpeg, datasetmisalignment, realtimedeepfake}. Convincing evidence is provided that these compression biases can be picked up by detectors, leading to high performance that does not reflect true generalization capabilities. Additionally, \citet{fakeorjpeg} and \citet{datasetmisalignment} also evaluate the impact of mismatches in image resolution between real and AI-generated images on detection performance. Because most detection models resize inputs to a fixed resolution, differences in the original image size introduce implicit upsampling or downsampling artifacts. When such resolution inconsistencies exist, detectors may rely on these artifacts rather than meaningful semantic cues, thereby introducing another source of bias that can significantly impact evaluation results. Lastly, differences in semantic content between AI-generated images and real images can create another bias detectors might rely on. \citet{imagebias} address this by proposing a data generation pipeline that minimizes such biases. Their approach generates images through self-conditioned reconstructions of real ones, producing samples that closely match the original semantic content. \textbf{These studies focus on AI-generated images, while we investigate such biases in AI-generated video detection, where temporal sampling introduces additional sources of shortcut learning.}

\section{Sampling biases}
\label{sec:bias}

In this section, we examine sampling and preprocessing biases in motion-based AI-generated video detectors. Specifically, we identified such biases in three out of the four previously mentioned detectors (\emph{D3}, \emph{ReStraV} and \emph{NSG-VD}). We use the official codebases of the papers, as well as the same datasets and train/test splits. If changes are made to the code, these are explicitly mentioned.
\paragraph{D3}
\label{sec:d3}
\citet{D3} propose a technique to detect AI-generated videos by taking the standard deviation of the second order differences of feature encodings of adjacent frames. They argue that this metric is generally lower for AI-generated videos due to the difficulty of generators to simulate second-order dynamics correctly. Their main evaluation is done on the GenVideo \citep{demamba} dataset: AI-generated videos are sourced from 10 different generators, while real videos are collected from the MSR-VTT \citep{msrvtt} (Microsoft Research Video to Text) dataset. Surprisingly, the real videos in the GenVideo dataset are saved at 3 frames per second (fps), while the real videos from the original MSR-VTT dataset are between 25 and 30 fps. \citet{D3} sample frames at 8 fps over a fixed duration, which introduces a major sampling bias: real videos at 3 fps are upsampled to 8 fps, resulting in frames being duplicated (up to three consecutive duplicate frames per video). This artificially introduces peaks in second order differences, which D3 can exploit as their detection signal. An example of this sampling bias can be found in Appendix \ref{sec:appendix_d3}. Table \ref{tab:d3} compares D3’s performance on the original GenVideo dataset (real videos at 3 fps) with its performance on the version of GenVideo where real videos are replaced by their original 25–30 fps versions. We follow the original codebase and limit the number of videos per class to 1000. We observe an average drop in AUC of 0.1736, confirming the substantial effect this bias has on the performance of D3. We note that this observation has also been independently reported by \citet{likelihoods}.

\begin{table}[hbtp]
\centering
\setlength{\tabcolsep}{2pt}
\caption{\textbf{Impact of sampling bias on the performance of D3.} The first row (D3) reports AUC per model against real videos of 3 fps from the GenVideo dataset (biased). The second row (D3*) reports AUC per model with real videos replaced by their high-fps versions (unbiased). On average, AUC drops by 0.1736 using the unbiased dataset.}
\label{tab:d3}
\fontsize{9.0pt}{9pt}\selectfont
\begin{tabular}{@{}lcccccccccc>{\columncolor{gray!30}}c@{}}
\toprule
\textbf{} & \textbf{Crafter} & \textbf{Gen2} & \textbf{HotShot} & \textbf{Lavie} & \textbf{MSE} & \textbf{MV} & \textbf{MSO} & \textbf{Show-1} & \textbf{Sora} & \textbf{WS} & \textbf{AVG} \\
\midrule
D3  & 0.9838 & 0.9926 & 0.9747 & 0.9727 & 0.9586 & 0.9907 & 0.9758 & 0.9847 & 0.9790 & 0.9178 & 0.9730\\
D3*  & 0.8438 & 0.9219 & 0.7775 & 0.7868 & 0.7291 & 0.9030 & 0.7693 & 0.8415 & 0.8088 & 0.6121 & 0.7994\\
\midrule
AUC $\downarrow$ & \textbf{0.1400} & \textbf{0.0707} & \textbf{0.1972} & \textbf{0.1859} & \textbf{0.2295} & \textbf{0.0877} & \textbf{0.2065} & \textbf{0.1432} & \textbf{0.1702} & \textbf{0.3057} & \textbf{0.1736}\\
\bottomrule
\end{tabular}

\end{table}

\paragraph{ReStraV}
\label{sec:restrav}

\citet{restrav} base their AI-generated video detector on the perceptual straightening hypothesis \citep{perceptual_straightening}, which states that the human visual system transforms complex motion into a straighter internal representation. They use DINOv2 \citep{DINOv2} to extract features from each frame and train an MLP on the first 7 distances and 6 angles (and the mean, variance, minimum and maximum of all distances and angles) between feature encodings of adjacent frames, resulting in 21 features. They claim that real videos follow a smoother path in the embedding space of DINOv2, a pattern that the MLP is able to identify. Their main evaluation is done on the VidProM \citep{vidprom} dataset, where 24 frames are sampled over a 2-second interval from each video. A bias is introduced in the way videos are handled that are less than 2 seconds long or have a lower fps than 12. For videos with fps lower than 12, frames are duplicated to reach this rate (similar to D3). Additionally, if a video is shorter than 2 seconds, the final frame is repeated until the required number of frames is obtained. We calculated one or both of these conditions to be true in 18715 out of 24997 AI-generated videos from the test split of the VidProM dataset used by ReStraV, while only 2 out of 24999 real videos trigger this behaviour.

This sampling bias influences the 21 features that the MLP is trained on. The most noticeable change is that the minimum distance value will be 0 if duplicate frames are present. We hypothesize that this is a pattern that can easily be exploited by the MLP to learn its decision boundary. To test this hypothesis, we simply use this one feature as our prediction value (without using the MLP) and compare the results to the original results in Table \ref{tab:restrav_single_feature}. We observe that for the generative models where sampling biases occur (Cog, MS, st2v, t2vz and vc2), performance is close to identical to the original ReStraV model, with an average AUC drop over the entire test split of the VidProM dataset of only 0.0535. Despite these major sampling biases, the original ReStraV model still achieves an AUC of 0.9355 on the unbiased Opensora videos. This indicates that even without sampling biases, there still exists a signal that the model can pick up to differentiate between AI-generated videos and real videos. As we will later show in Section \ref{sec:motion}, a large part of this performance can be attributed to motion biases in the dataset.

\begin{table}[htbp]
\centering
\caption{\textbf{Analysis of sampling biases in the ReStraV detector.} The first row shows the total number of videos per model in the VidProM test set. The second row shows the number of videos per model in the VidProM test set where sampling biases occur. The third row shows AUC results of the original ReStraV model, while the fourth row shows AUC results using the minimum distance value between consecutive encodings directly as the prediction. This single metric achieves on average only a 0.0535 drop in AUC compared to the original ReStraV detector.}
\label{tab:restrav_single_feature}
\fontsize{9.0pt}{9pt}\selectfont
\begin{tabular}{@{}lccccccc>{\columncolor{gray!30}}c@{}}
\toprule
\textbf{} & \textbf{Cog} & \textbf{MS} & \textbf{Opensora} & \textbf{Pika} & \textbf{st2v} & \textbf{t2vz} & \textbf{vc2} & \textbf{AVG} \\
\midrule
Total number & 3551 & 3574 & 3608 & 3583 & 3630 & 3581 & 3470 \\
Biased number & 3551 & 3574 & 0 & 909 & 3630 & 3581 & 3470 \\
\midrule
ReStraV  & 0.9983 & 1.0000 & 0.9355 & 0.9769 & 0.9912 & 1.0000 & 0.9997 & 0.9859\\
ReStraV min. dist. & 0.9920 & 0.9920 & 0.6846 & 0.8825 & 0.9920 & 0.9920 & 0.9920 & 0.9324 \\ 
\midrule
AUC $\downarrow$ & 0.0062 & 0.0079 & 0.2509 & 0.0944 & -0.0009 & 0.0079 & 0.0076 & 0.0535 \\
\bottomrule
\end{tabular}

\end{table}

\begin{figure}[htbp]
    \centering
   
    \includegraphics[width=\linewidth]{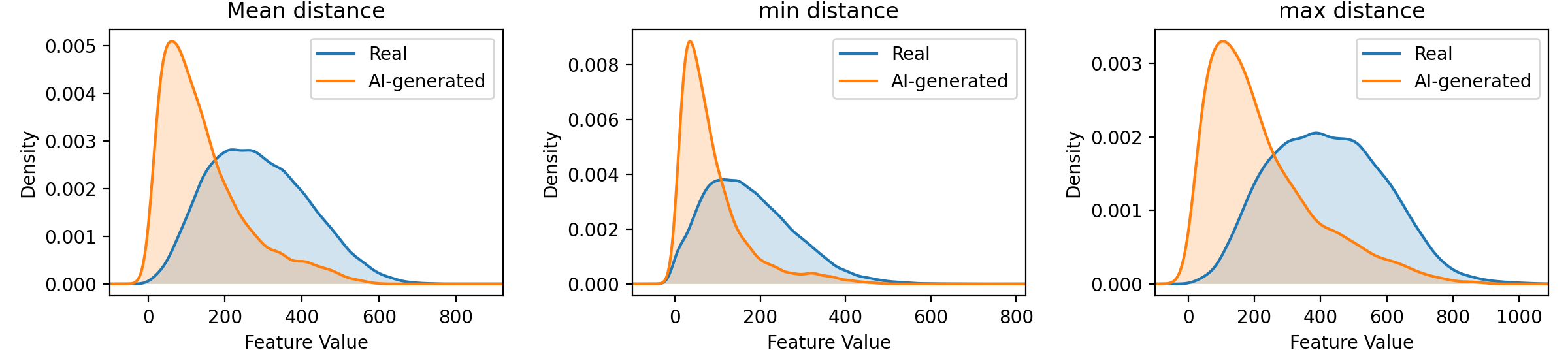}  
     \caption{Distributions of the mean, minimum and maximum distance values of all unbiased samples in the test set of VidProM. AI-generated videos tend to have lower distance values than real videos, which contradicts the theoretical assumptions and evaluations of the original ReStraV paper.}
        \label{fig:restrav_features}
\end{figure}

In addition to these sampling biases, we observed a fundamental difference between the results and the theoretical assumptions of ReStraV. Their claim that real videos have a smoother trajectory in the representation space of a pre-trained vision encoder contradicts the conclusions of other papers in the field that use similar approaches. \citet{D3}, for example, report that \emph{"real videos exhibit more chaotic speed variations"}, while \emph{"generated videos tend to show very flat patterns"}. Accordingly, \citet{Over-Coherence} come to the conclusion that \emph{"unnaturally high temporal coherence is a key indicator of generated text-to-video content"}. To analyze this discrepancy, we recreated part of Figure 5 in the ReStraV paper by plotting the distribution of the mean, minimum and maximum distance values of the unbiased subset of the VidProM test set, which can be seen in Figure \ref{fig:restrav_features}. To our surprise, we get almost completely opposite distributions than those that are presented in the original paper, with AI-generated videos having lower distances on average than real videos. These results actually align with the conclusions of \citet{D3} and \citet{Over-Coherence}, and with our motion bias analysis in Section \ref{sec:motion}. While this substantial difference doesn’t automatically invalidate ReStraV’s findings, it does raise significant concerns about the validity of its theoretical foundation.

\paragraph{NSG-VD}
\label{sec:nsgvd}
\citet{NSGVD} propose an AI-generated video detection framework based on \emph{Normalized Spatiotemporal Gradients (NSG)}. This metric combines the output of the score function of a pretrained diffusion model together with inter-frame pixel-wise differences and is based on the physics theory of probability flow conservation: 
\begin{equation}
\label{eq:nsgvd}
    \mathbf{g}(\mathbf{x}, t) \approx \frac{\mathbf{s}_\theta(\mathbf{x}_t)}{\mathbf{s}_\theta(\mathbf{x}_t) \cdot \frac{\mathbf{x}_{t+\Delta t} - \mathbf{x}_t}{\Delta t} + \lambda} \quad \text{\citep{NSGVD}}
\end{equation} 
They use Maximum Mean Discrepancy (MMD) to measure these NSG features against the NSG features of a reference set of real videos. They train the kernel used by MMD on a train set of real videos from the Kinetics-400 dataset \citep{Kinetics} and AI-generated videos from either SEINE or Pika, sourced from the GenVideo train set \citep{demamba}. They perform their main evaluation on the GenVideo validation set, similar to D3 \citep{D3}. Sampling is performed by extracting eight frames uniformly spaced across the entire video sequence. Consequently, this sampling method establishes a correlation between the total video length and the temporal distance between consecutive sampled frames. In the GenVideo validation set, real videos have an average duration of 15.52 seconds, while AI-generated videos have an average duration of 2.98 seconds. This means that the sampling method of NSG-VD results in an average time between frames of 1.94 seconds for real videos compared to 0.3725 seconds for AI-generated videos; a more than 5-fold difference. 
Also, in the official codebase, the division by $\Delta t$ in Equation \ref{eq:nsgvd} is not performed, mathematically treating the temporal gap between adjacent frames as identical for all videos. 

\begin{figure}[h!]
    \centering
   
    \includegraphics[width=\linewidth]{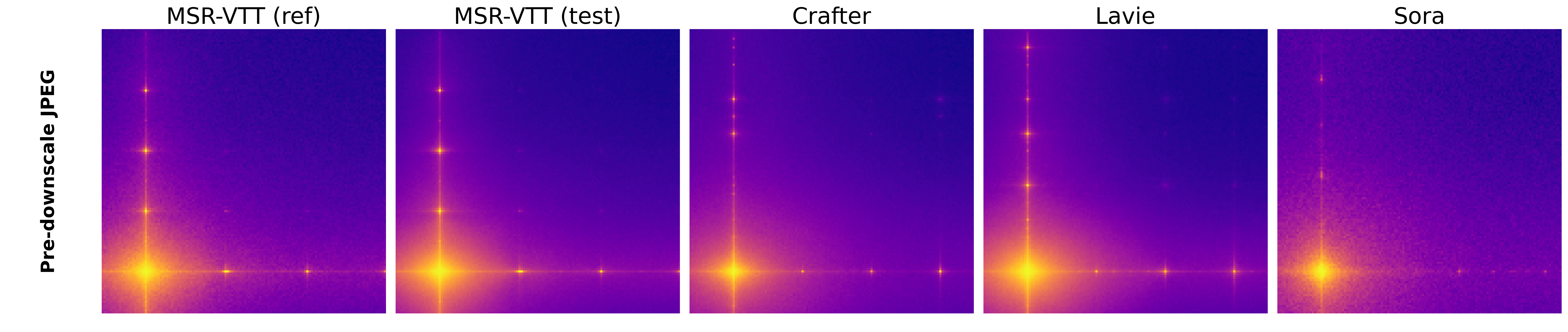}
    \includegraphics[width=\linewidth]{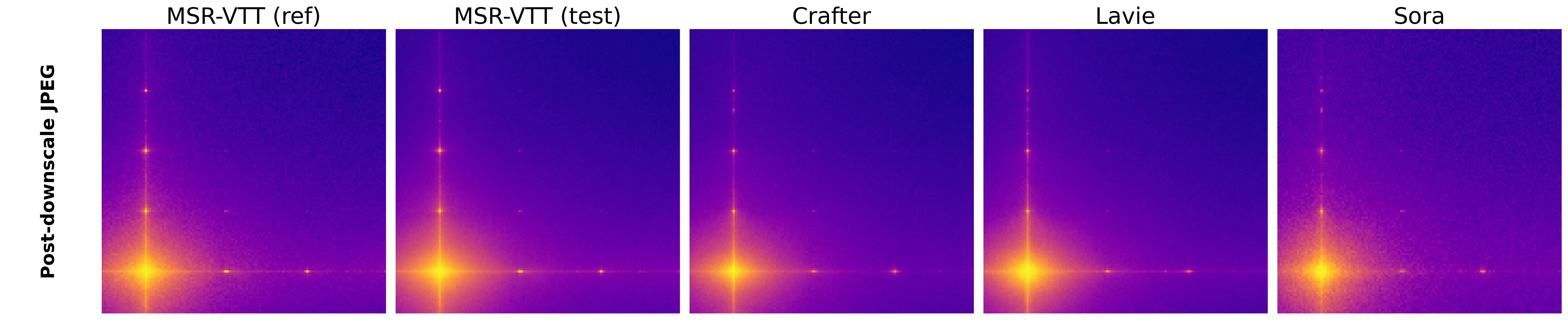}
     \caption{Average FFT representation (top-right quadrant) of the output of the score function used by the NSG-VD detector on the GenVideo dataset under different preprocessing pipelines. First row: JPEG compression before downscaling (original NSG-VD). Second row: JPEG compression after downscaling (at 298×224 resolution with a quality factor of 95). In the first row, the locations of the spectral peaks are different between the score output of AI-generated videos and the real reference set, while in the second row, the location of these peaks are identical. An extended version of this figure, including the output visualised in the spatial domain, is provided in Appendix \ref{sec:appendix_nsgvd}.}
    \label{fig:score}
\end{figure}
\begin{table}[b!]
\centering
\caption{Results for the NSG-VD detector on the GenVideo dataset under different preprocessing pipelines. (1) JPEG compression before downscaling (original NSG-VD), (2) No JPEG compression and (3) JPEG compression after downscaling. An AUC drop of 0.2794 is observed between the post-downscale and pre-downscale JPEG encoding pipelines.}
\label{tab:nsgvd}
\resizebox{\textwidth}{!}{%
\begin{tabular}{@{}lcccccccccc>{\columncolor{gray!30}}c@{}}
\toprule
\textbf{} & \textbf{Crafter} & \textbf{Gen2} & \textbf{HotShot} & \textbf{Lavie} & \textbf{MSE} & \textbf{MV} & \textbf{MSO} & \textbf{Show-1} & \textbf{Sora} & \textbf{WS} & \textbf{AVG} \\
\midrule
(1) Pre-downscale JPEG  & 0.9927 & 0.9873 & 0.8979 & 0.9468 & 0.9525 & 0.9990 & 0.9836 & 0.9554 & 0.9632 & 0.9640 & 0.9642 \\
(2) No JPEG                  & 0.9861 & 0.9763 & 0.9385 & 0.9579 & 0.9390 & 0.9934 & 0.9756 & 0.9628 & 0.9961 & 0.9511 & 0.9677 \\
(3) Post-downscale JPEG & 0.7411 & 0.6276 & 0.7030 & 0.6772 & 0.6473 & 0.7755 & 0.6713 & 0.7312 & 0.5714 & 0.7024 & 0.6848 \\
\bottomrule
\end{tabular}%
} 
\end{table}

A second bias is introduced in the way the score function behaves relative to the resolution of the original video. \citet{NSGVD} first extract 8 frames from each H.264-encoded video at their native resolution and save them in JPEG format. These frames are subsequently loaded and resized to 256x256 pixels to serve as input for the diffusion model. This pipeline introduces a resolution-dependent distortion: when high-resolution videos are significantly downscaled, the JPEG and H.264 quantization blocks are suppressed. Conversely, when a video's native resolution is already close to 256x256, minimal scaling occurs, and these grid artifacts remain intact. All real videos in the GenVideo dataset have a resolution of 298x224, close to the input resolution of the diffusion model. AI-generated videos in the GenVideo dataset have resolutions ranging from 512x320 (Lavie) to 1920x1080 (Sora). Because the MMD kernel compares these scores against a reference set of real videos, it may inadvertently detect these resolution-linked JPEG/H.264 signatures rather than the underlying generative quality of the video. To analyze the effect of this resolution bias, we evaluate three preprocessing variations: (1) JPEG compression before downscaling (original NSG-VD), (2) No JPEG compression and (3) JPEG compression after downscaling (at 298×224 resolution with a quality factor of 95). For pipeline (1) and (3), we visualize the average FFT transform of the output of the score function of NSG-VD for the real reference set, the real test set and three different models from the GenVideo dataset in Figure \ref{fig:score}. Compression artifacts can be identified by specific spectral peaks on the horizontal and vertical axes in the FFT domain. Clear differences can be seen in the locations of these peaks using the original pipeline (1) of NSG-VD between AI-generated videos and the real reference set. This is a direct consequence of the differences in resolution between both classes. However, when JPEG compression is applied after downscaling (3), the spectral peaks become indistinguishable between real and generated videos. Table \ref{tab:nsgvd} shows a substantial performance drop of 0.2794 AUC under this setting. This indicates that NSG-VD largely relies on resolution-related artifacts present in the preprocessing pipeline, rather than learning true discriminative features between real and AI-generated videos.

\section {Motion biases}
\label{sec:motion} 

We perform our motion bias analysis on five recent AI-generated video datasets and evaluate the impact of these biases for three motion-based detectors: \emph{D3}, \emph{ReStraV} and \emph{Over-Coherence}. Four of the datasets are the main evaluation datasets of the detectors we evaluate: \emph{GenVideo} \citep{demamba} for D3, \emph{VidProM} \citep{vidprom} for ReStraV, and the datasets introduced alongside \emph{Over-Coherence} \citep{Over-Coherence} and \emph{WaveRep} \citep{waverep}. As a fifth dataset we include \emph{GenBuster++} \citep{buster}, which, as we show later in this Section, does not exhibit the motion bias present in the other four.

\paragraph{Evaluation setup}
For the \emph{GenVideo} dataset, we use the high-fps versions of real videos, as explained in Section \ref{sec:d3}. We also remove the WildScrape subset due to a lack of quality samples. For the \emph{VidProM} dataset, we use the test split as used in the ReStraV codebase. To ensure a fair comparison, we use the same frame sampling pipeline for each dataset and detector. It is important to note that this pipeline differs from the individual pipelines employed for each detector. Although this might result in different absolute evaluation metrics, we argue that it is essential for a controlled evaluation of motion-related biases. Each video is sampled at 8 frames per second over a 2-second interval. Videos that have a frame rate lower than 8 or that are shorter than 2 seconds are discarded. The original and discarded number of videos can be found in Appendix \ref{app:dataset}, as well as the generators used in each dataset. Frames are saved in the lossless PNG format. Each frame is resized so that its shortest side is 224 pixels, followed by a center crop to a resolution of 224×224. All images are converted to RGB and normalized depending on the normalization used in the original codebases (ImageNet \citep{ImageNet} normalization for D3, CLIP \citep{clip} normalization for Over-Coherence and standard [0, 1] normalization for ReStraV). 

For D3, we use the original implementation without modification. For ReStraV, because it was initially trained with a biased sampling strategy (as discussed in Section \ref{sec:restrav}), we retrain the MLP using our preprocessing pipeline as described above. We keep the same train split as in the original setup, but exclude unsatisfactory videos as previously outlined in our evaluation setup. For Over-Coherence, the method relies on a threshold T to select either the minimum or maximum cosine similarity between consecutive frame embeddings. The original paper reports using T=0.99, while the official implementation uses T=0.995. In our experiments, we find that T=1.0 (always selecting the minimum cosine similarity) gives the best performance under our preprocessing pipeline. Given that \citet{Over-Coherence} report only a 0.01 AUC drop when only using the minimum cosine similarity, we consider this adjustment justified for our setting.

\paragraph{Dataset motion biases}
\label{sec:4.2}

\begin{figure}[h!]
    \centering
    
    \includegraphics[width=\linewidth]{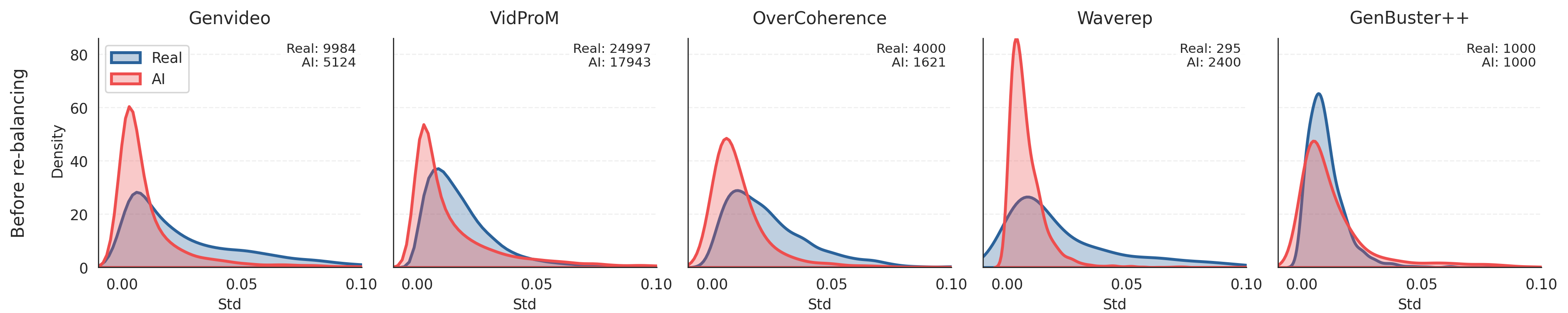}
    \caption{Distribution of the motion std metric for all evaluated datasets. In the GenVideo, VidProM, Over-Coherence and WaveRep datasets, real videos tend to have more erratic motion patterns, while AI-generated videos have smoother motion patterns. In the Genbuster++ dataset, this distribution is more balanced. The distributions and number of videos after rebalancing are shown in Appendix \ref{sec:appendix_rebalanced}.}
        \label{fig:motion_distributions}

\end{figure}

We hypothesize that the aforementioned datasets contain real videos that, in general, contain more erratic motion and scene transitions than AI-generated videos. To demonstrate this bias, we use a very simple metric: the standard deviation of the mean absolute pixel difference between consecutive frames of a video. This metric measures the temporal irregularity of frame-to-frame change, and therefore responds to scene cuts, camera and object motion, and lighting changes. We deliberately choose such a simple statistic, since if a single non-learned temporal statistic can separate real and AI-generated videos, the benchmarks contain a label-correlated temporal shortcut. Qualitative examples (Appendix \ref{sec:appendix_motion_bias_examples}) show that in these datasets this metric correlates with scene cuts and camera and object motion. We plot the distributions of this metric for each dataset in Figure \ref{fig:motion_distributions}. As can be seen, for each dataset, the distribution of this metric is not equal. In GenVideo, WaveRep, Over-Coherence, and VidProM, AI-generated videos tend to show less temporal variation than real videos. The only exception is GenBuster++, where AI-generated videos display a standard deviation similar to that of real videos. The GenBuster++ dataset is specifically designed to have a strong alignment with real-world scenarios \citep{buster}. They achieve this by extracting prompts from real videos and using those to generate AI-videos. They then give all videos (real and generated) blindly to a group of experts and only keep the generated videos that were identified as real by the experts. This pipeline explains our results and motivates the fact that the motion differences in the other four datasets are not inherent to AI-generated videos, but rather a bias stemming from unaligned data and possibly outdated models. To analyze the effect this dataset imbalance has on the results of the aforementioned models, we apply a re-balancing strategy. We partition all videos into 100 bins based on the previously explained standard deviation metric. Within each bin, we randomly undersample the majority class to match the minority class. This results in a filtered dataset in which real and AI-generated videos share a similar distribution of this metric.

\paragraph{Evaluation of detectors}
 We evaluate the models on the original datasets and the rebalanced datasets. We compute the AUC and report the results in Table \ref{tab:model_evaluation}. 
 The first notable finding is that all three detectors perform better than random on the four biased datasets, but drop to random-level performance on the GenBuster++ dataset, which does not contain the aforementioned motion bias. This result alone strongly indicates that the signals these detectors rely on are not inherent properties of AI-generated videos, but are instead a reflection of motion imbalances present in their evaluation datasets. Remarkably, our custom motion metric achieves second or close-to-second performance on all datasets. While we do not claim to have developed a new competitive detector, the fact that such a simple metric can rival state-of-the-art detectors implies that these detectors are largely using the same trivial motion distribution signal rather than features that are genuinely discriminative of AI-generated videos.

This conclusion is further supported by the rebalancing experiments. On the rebalanced datasets (GenBuster++ excluded), all detectors experience a performance drop in AUC (D3: -0.0970, ReStraV: -0.0650, Over-Coherence: -0.1127). Additionally, to show the brittleness of only relying on inter-frame encoding differences for detection, we apply a Gaussian blur of 7x7 to a single random frame in all AI-generated videos. This augmentation results in two consecutive encodings being further away from each other, while having a minimally noticeable impact on the original video. As can be seen in Table \ref{tab:model_evaluation}, this simple attack causes extreme AUC drops across all datasets (excluding GenBuster++) for all detectors evaluated (D3: -0.3795, ReStraV: -0.2202, Over-Coherence: -0.5693), either collapsing to random predictions or even dropping below an AUC of 0.5. We show the results for this specific augmentation, but suspect a wide range of augmentation attacks can achieve a similar effect. For detectors to be useful in real-world scenarios, simple augmentations like the one we evaluated should not have this big of an impact.

\begin{table}[h!]
\centering
\caption{AUC results for all datasets and models evaluated using our unified preprocessing pipeline. \textbf{Original:} results using the full dataset. \textbf{Rebalanced:} results using the rebalanced datasets. \textbf{Blur frame (attack):} results where a random frame of each AI-generated video is blurred.}
\label{tab:model_evaluation}
\fontsize{9.0pt}{9pt}\selectfont
\begin{tabular}{llcccccc}
\toprule

\textbf{Dataset} & \textbf{Model} 
& \textbf{Original} 
& \multicolumn{2}{c}{\textbf{Rebalanced}} 
& \multicolumn{2}{c}{\textbf{Blur frame (attack)}} \\
\cmidrule(lr){4-5} \cmidrule(lr){6-7}

& & \textbf{AUC} 
& \textbf{AUC} & \textbf{$\Delta$} 
& \textbf{AUC} & \textbf{$\Delta$} \\
\midrule

GenVideo
    & D3             & \textbf{0.7941} & \textbf{0.6728} & -0.1213 & 0.4563 & -0.3378 \\
    & ReStraV        & 0.7130 & \underline{0.6448} & -0.0682 & \underline{0.4736} & -0.2394 \\
    & Over-Coherence & 0.7221 & 0.5561 & -0.1660 & 0.2532 & -0.4689 \\
    & Motion std metric    & \underline{0.7224} & 0.5024 & -0.2200 & \textbf{0.6628} & -0.0596 \\

\midrule

VidProM
    & D3             & 0.6103 & \underline{0.5466} & -0.0637 & 0.2670 & -0.3433 \\
    & ReStraV        & \textbf{0.8488} & \textbf{0.8028} & -0.0460 & \textbf{0.7037} & -0.1451 \\
    & Over-Coherence & 0.5599 & 0.4745 & -0.0854 & 0.0575 & -0.5024 \\
    & Motion std metric   & \underline{0.6342} & 0.5072 & -0.1270 & \underline{0.5861} & -0.0481 \\

\midrule

Over-Coherence
    & D3             & \underline{0.7546} & \underline{0.6475} & -0.1071 & 0.3343 & -0.4203 \\
    & ReStraV        & 0.6332 & 0.5443 & -0.0889 & \underline{0.4068} & -0.2264 \\
    & Over-Coherence & \textbf{0.8162} & \textbf{0.7314} & -0.0848 & 0.1090 & -0.7072 \\
    & Motion std metric    & 0.7395 & 0.5024 & -0.2371 & \textbf{0.7185} & -0.0210 \\

\midrule

WaveRep
    & D3             & \textbf{0.7190} & \underline{0.6230} & -0.0960 & 0.3026 & -0.4164 \\
    & ReStraV        & \underline{0.7187} & \textbf{0.6618} & -0.0569 & \underline{0.4487} & -0.2700 \\
    & Over-Coherence & 0.7089 & 0.5944 & -0.1145 & 0.1102 & -0.5987 \\
    & Motion std metric   & 0.7103 & 0.4975 & -0.2128 & \textbf{0.6596} & -0.0507 \\

\midrule

GenBuster++
    & D3             & 0.4843 & 0.4955 & 0.0112 & 0.1478 & -0.3365 \\
    & ReStraV        & \underline{0.5154} & 0.4934 & -0.0220 & \underline{0.3252} & -0.1902 \\
    & Over-Coherence & \textbf{0.5434} & \textbf{0.5562} & 0.0128 & 0.0225 & -0.5209 \\
    & Motion std metric   & 0.5023 & \underline{0.5006} & -0.0017 & \textbf{0.4440} & -0.0583 \\

\bottomrule
\end{tabular}
\end{table}

\section{Conclusions}
\label{sec:conclusions}
\paragraph{Frequency-based detectors}
The above results raise the question of whether the observed limitations are inherent to video detection, or specific to motion-based approaches. To answer this question, we evaluate an additional detector on the five datasets used in our main evaluation. We chose the WaveRep detector \citep{waverep}, a recent approach that trains a DinoV2 model using an augmentation strategy that replaces specific low-frequency bands in AI-generated videos with real counterparts to force the model to focus on mid- to high-frequency artifacts. It is important to note that WaveRep does frame-by-frame prediction and averages those predictions, thus inherently not exploiting motion artifacts. WaveRep achieves consistently strong performance across all evaluated datasets, as can be seen in Table \ref{tab:waverep_evaluation}. While a robustness analysis of frequency-based detectors is beyond the scope of this work, these results suggest that the limitations we identified may be specific to motion-based approaches rather than inherent to AI-generated video detection more broadly. 
We hope these results motivate future work to investigate what signals frequency-based detectors exploit, and whether they are similarly vulnerable to targeted adversarial attacks and distribution shifts.

\begin{table}[htbp]
\centering
\caption{AUC results for the WaveRep detector on all datasets. The evaluation follows the same protocol as Table \ref{tab:model_evaluation}. Frames are read at a resolution of 224x224 and upscaled afterwards to 504x504 to match the input resolution of the WaveRep detector.}
\label{tab:waverep_evaluation}
\fontsize{9.0pt}{9pt}\selectfont
\begin{tabular}{llcccccc}
\toprule
\textbf{Dataset} & \textbf{Model} 
& \textbf{Original} 
& \multicolumn{2}{c}{\textbf{Rebalanced}} 
& \multicolumn{2}{c}{\textbf{Blur frame (attack)}} \\
\cmidrule(lr){4-5} \cmidrule(lr){6-7}

& & \textbf{AUC} 
& \textbf{AUC} & \textbf{$\Delta$} 
& \textbf{AUC} & \textbf{$\Delta$} \\

\midrule

GenVideo
    & WaveRep & 0.9893 & 0.9895 & 0.0002 & 0.9889 & -0.0004 \\

VidProM
    & WaveRep & 0.9795 & 0.9777 & -0.0018 & 0.9786 & -0.0009 \\

Over-Coherence
    & WaveRep & 0.9892 & 0.9904 & 0.0012 & 0.9889 & -0.0003 \\

WaveRep
    & WaveRep & 0.9950 & 0.9963 & 0.0013 & 0.9950 & 0.0000 \\

GenBuster++
    & WaveRep & 0.9191 & 0.9107 & -0.0084 & 0.9144 & -0.0047 \\

\bottomrule
\end{tabular}
\end{table}

\paragraph{Implications for future datasets and detectors}

We showed in this paper that the high accuracies reported in state-of-the-art AI-generated video detectors do not translate well to real applicability outside their evaluation setup. Our findings point to several practical considerations for future datasets, detectors and evaluation setups. Firstly, great care should be taken that real and AI-generated videos have similar characteristics (duration, fps, resolution...). If there is a substantial difference in these characteristics, they should be explicitly mentioned and correctly handled. As we showed in Section \ref{sec:bias}, if differences in fps and resolution are not accounted for, detectors can pick up these biases instead of detecting real discriminative features between AI-generated and real videos. 

Next, the artifacts that a detector relies on should be representative of the current state of generative models, rather than tied to specific biases in evaluation datasets. As generative models rapidly improve, signals that were once discriminative, such as limited motion in AI-generated videos, may quickly become obsolete, as we have shown in this paper. Finally, a proposed detector should be compared against a trivial non-learned baseline. If a single feature can rival a state-of-the-art method, as the minimum distance value in Section \ref{sec:restrav} and the motion metric in Section \ref{sec:4.2} do, the reported performance likely reflects a dataset-level property rather than a discriminative one. 

These recommendations reduce the risk of shortcut learning, but they do not eliminate it. Our own rebalancing corrects the specific imbalance we measured, yet this does not make the resulting datasets free of all (motion-)related biases. More broadly, we do not believe it is possible to guarantee a dataset is completely free of bias, since the exact biases a model might exploit can never be fully known in advance. We see identifying and mitigating specific, concrete biases (ours and prior work in Section \ref{sec:related_work}) as valuable precisely because it narrows this gap, even if it cannot close it.


\subsubsection*{Acknowledgments}
This research was funded by the FWO fellowship grant (1S81026N) and the NORM.AI SBO
project, funded by Flanders Make, the strategic research center for the manufacturing industry in
Belgium. This work was made possible with support from MAXVR-INFRA, a scalable and flexible
infrastructure that facilitates the transition to digital-physical work environments.

\bibliography{iclr2027_conference}
\bibliographystyle{iclr2027_conference}

\appendix
\clearpage
\section{Appendix}
\subsection{Datasets overview}
\label{app:dataset}
We report the number of videos in the five datasets we evaluated in Table \ref{tab:datasets_preprocessing}. We also report the number of videos that remain after unsatisfactory videos are filtered by our preprocessing pipeline. In Table \ref{tab:dataset_generators}, we report the video generators included in each dataset evaluated in Section \ref{sec:motion}. Most
generators appear in at most two datasets, indicating that the datasets are largely complementary in
terms of generator coverage.
\begin{table}[h!]
\centering
\caption{Overview of the datasets used for our motion bias evaluation. Videos are filtered if they have an fps lower than 8 or if they are less than 2 seconds long.}
\label{tab:datasets_preprocessing}
\resizebox{\linewidth}{!}{
\begin{tabular}{lcccc}
\toprule
\textbf{Dataset} & \textbf{AI-generated} & \textbf{Real} & \textbf{Filtered AI-generated} & \textbf{Filtered Real} \\
\midrule
GenVideo \citep{demamba} & 7660 & 10000 & 5124 & 9984 \\
VidProM \citep{vidprom} & 24997 & 24999 & 17943 & 24997 \\
Over-Coherence \citep{Over-Coherence} & 1680 & 4000 & 1621 & 4000 \\
WaveRep \citep{waverep} & 2400 & 295 & 2400 & 295 \\
GenBuster++ \citep{buster} & 1000 & 1000 & 1000 & 1000 \\
\bottomrule
\end{tabular}}
\end{table}

\begin{table}[h!]
    \centering
    \renewcommand{\arraystretch}{1.3}
    \setlength{\tabcolsep}{8pt}
    \caption{Generators used for the AI-generated videos in each of the five evaluated datasets.}
    \label{tab:dataset_generators}
    \resizebox{\linewidth}{!}{
    \begin{tabular}{l>{\raggedright\arraybackslash}p{0.68\linewidth}}
        \toprule
        \textbf{Dataset} & \textbf{Generators} \\
        \midrule
        GenVideo \citep{demamba} & Crafter, Gen2, HotShot, Lavie, MorphStudio, MoonValley, ModelScope, Show-1, Sora \\
        \addlinespace[4pt]
        VidProM \citep{vidprom} & CogVideo, ModelScope, OpenSora, Pika, Text2Video-Zero, VideoCrafter2, StreamingT2V \\
        \addlinespace[4pt]
        Over-Coherence \citep{Over-Coherence} & LTX-Video, HunyuanVideo, Wan-2.1 \\
        \addlinespace[4pt]
        WaveRep \citep{waverep} & Allegro, CogVideo, Mochi-1, OpenSora-Plan, Sora, NOVA, FLUX \\
        \addlinespace[4pt]
        GenBuster++ \citep{buster} & Wan2.1, Veo 3, Veo 2, Sora, SkyReels V1, Seedance 1.0, PixVerse V4, Pika, Luma Ray2, Kling 2.0, HunyuanVideo, Hailuo Video 01, Runway Gen-3 \\
        \bottomrule
    \end{tabular}}
\end{table}

\clearpage

\subsection{Rebalanced datasets}
\label{sec:appendix_rebalanced}
\begin{figure}[h!]
    \centering
    
    \includegraphics[width=\linewidth]{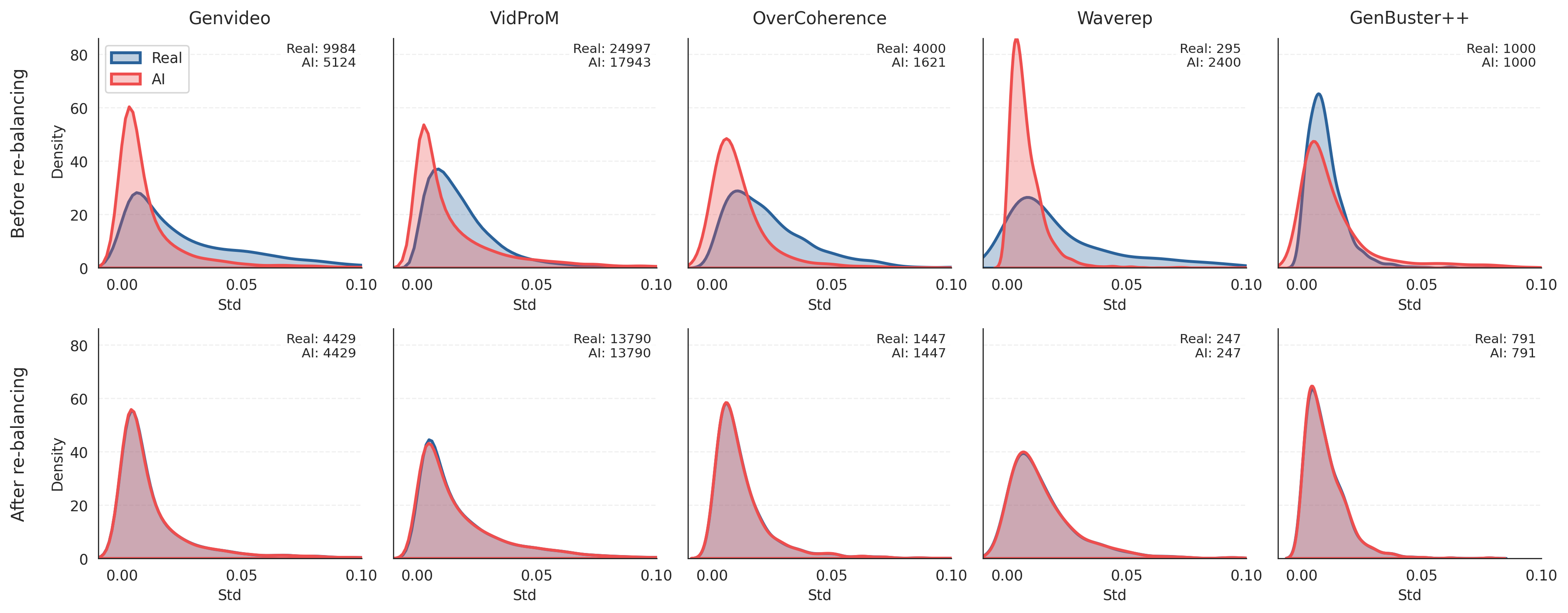}
    \caption{Distribution of the motion std metric for all evaluated datasets. In the GenVideo, VidProM, Over-Coherence and Waverep datasets, real videos tend to have more erratic motion patterns, while AI-generated videos have smoother motion patterns. In the Genbuster++ dataset, this distribution is more balanced. The bottom row shows the distributions after rebalancing, as well as the number of AI-generated and real videos that remain.}
        \label{fig:motion_distributions_full} 
\end{figure}

We extend Figure \ref{fig:motion_distributions} of the main paper with the distributions and number of videos after rebalancing in Figure \ref{fig:motion_distributions_full}.

\subsection{D3 sampling bias example}
\label{sec:appendix_d3}

\begin{figure}[h!]
    \centering
    
    \includegraphics[width=\linewidth]{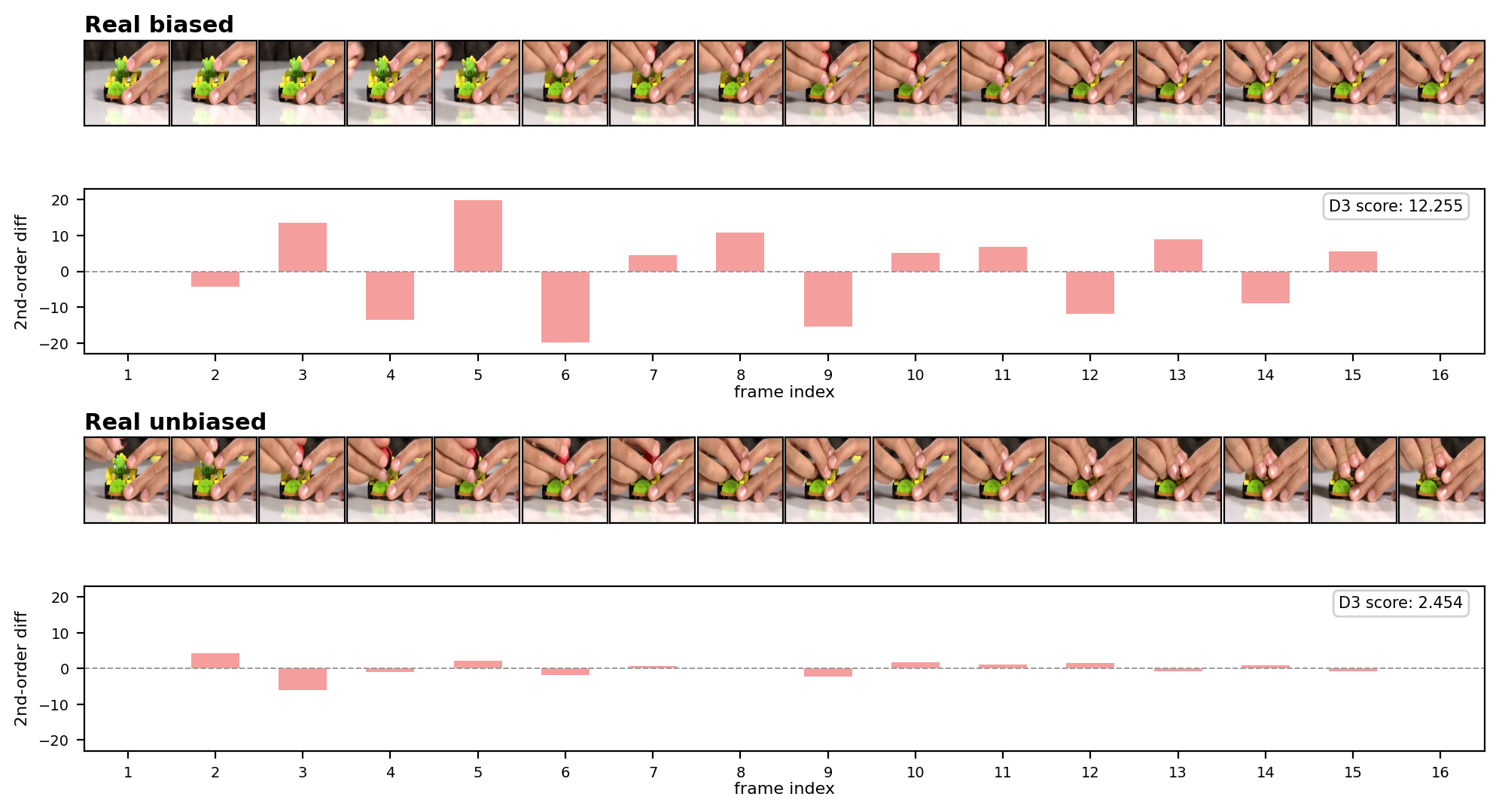}
    \caption{The top row shows a real video from
the GenVideo dataset, stored at 3 fps and upsampled to 8 fps following D3’s preprocessing pipeline.
The upsampling inserts duplicate frames, inflating the variance of the second-order embedding
differences on which D3 relies. The bottom row shows the same video sampled directly from the
source dataset (MSR-VTT) at 8 fps. Without duplicate frames, this variance drops sharply, and D3’s
prediction shifts toward the "AI-generated" class.}
        \label{fig:d3_failure_cases}

\end{figure}

Figure \ref{fig:d3_failure_cases} illustrates the sampling bias of D3, discussed in Section \ref{sec:d3}.

\clearpage
\subsection{Score function outputs of NSG-VD}
\label{sec:appendix_nsgvd}
\begin{figure}[h!]
    \centering
   
    \includegraphics[width=\linewidth]{Figures/Images/paper_figure_fft_nsg-vd-low-fps-jpg.png}
    \includegraphics[width=\linewidth]{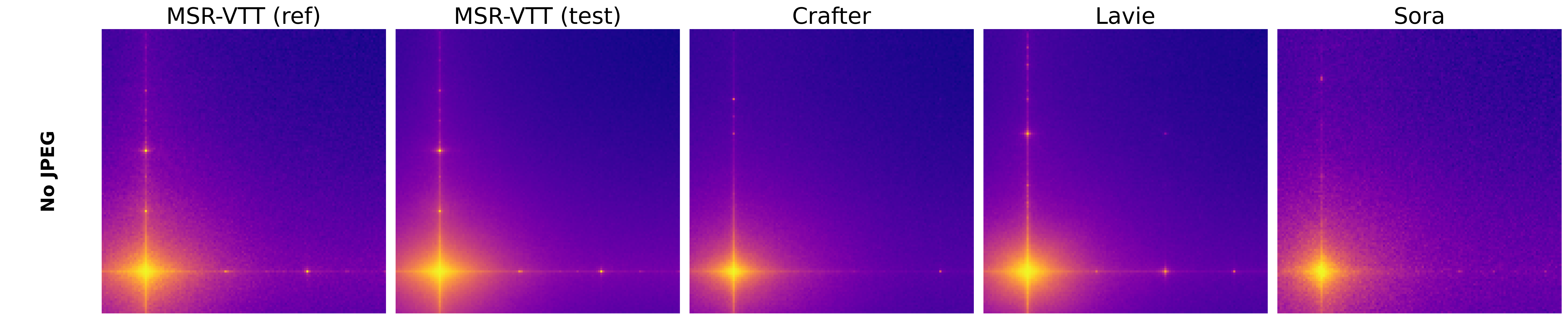}
    \includegraphics[width=\linewidth]{Figures/Images/paper_figure_fft_nsg-vd-low-fps-post-jpg.png}
    \includegraphics[width=\linewidth]{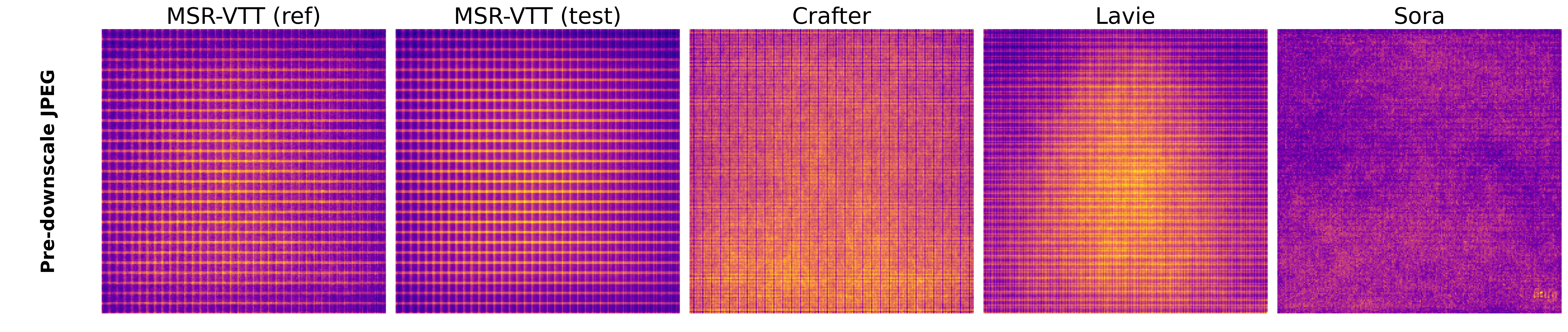}
    \includegraphics[width=\linewidth]{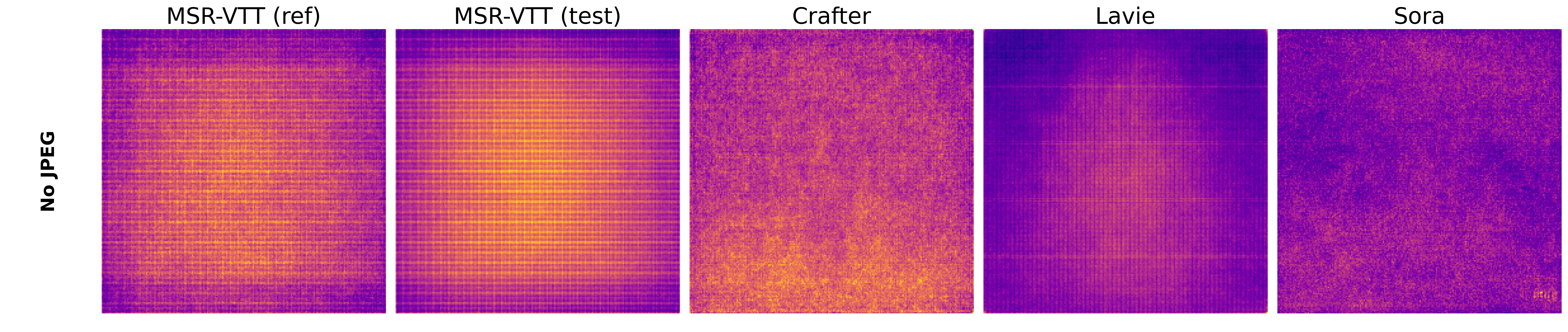}
    \includegraphics[width=\linewidth]{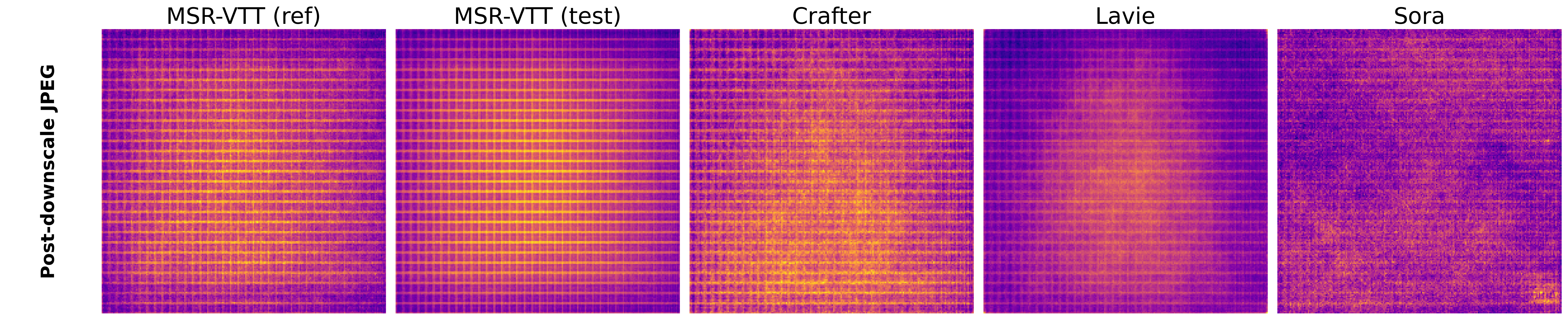}

     \caption{\textbf{Rows 1, 2 and 3}: Average FFT representation (top-right quadrant) of the output of the score function used by the NSG-VD detector on the GenVideo dataset under different preprocessing pipelines. First row: JPEG compression before downscaling (original NSG-VD). Second row: no JPEG compression. Third row: JPEG compression after downscaling (at 298×224 resolution with a quality factor of 95). In the first and second rows, the locations of the spectral peaks are different between the score output of AI-generated videos and the real reference set, while in the third row, the location of these peaks are identical. \textbf{Rows 4, 5 and 6}: the same setup, but in the spatial domain.}
    \label{fig:score_extended}
\end{figure}

\subsection{Motion bias examples}
\label{sec:appendix_motion_bias_examples}

\begin{figure}[h!]
    \centering
    
    \includegraphics[width=\linewidth]{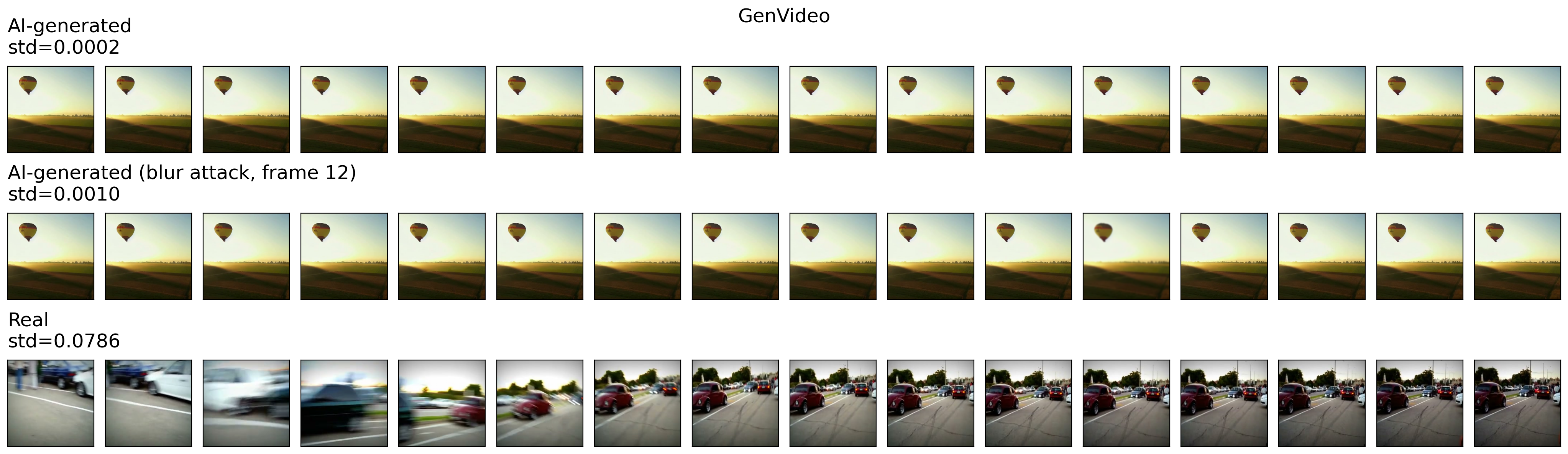}
    \includegraphics[width=\linewidth]{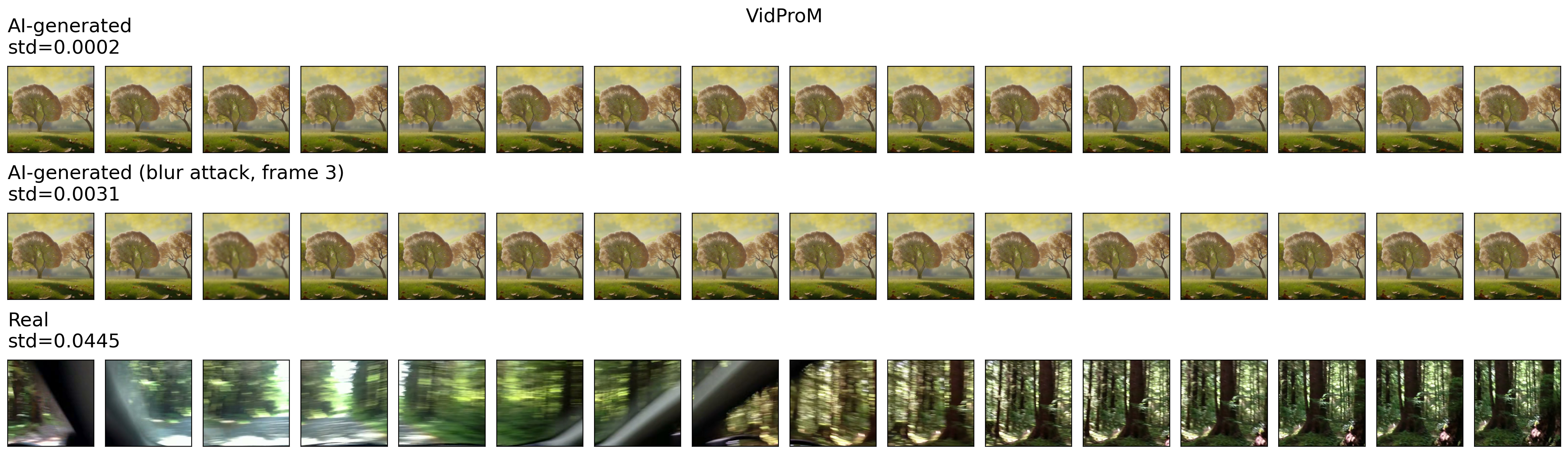}
    \includegraphics[width=\linewidth]{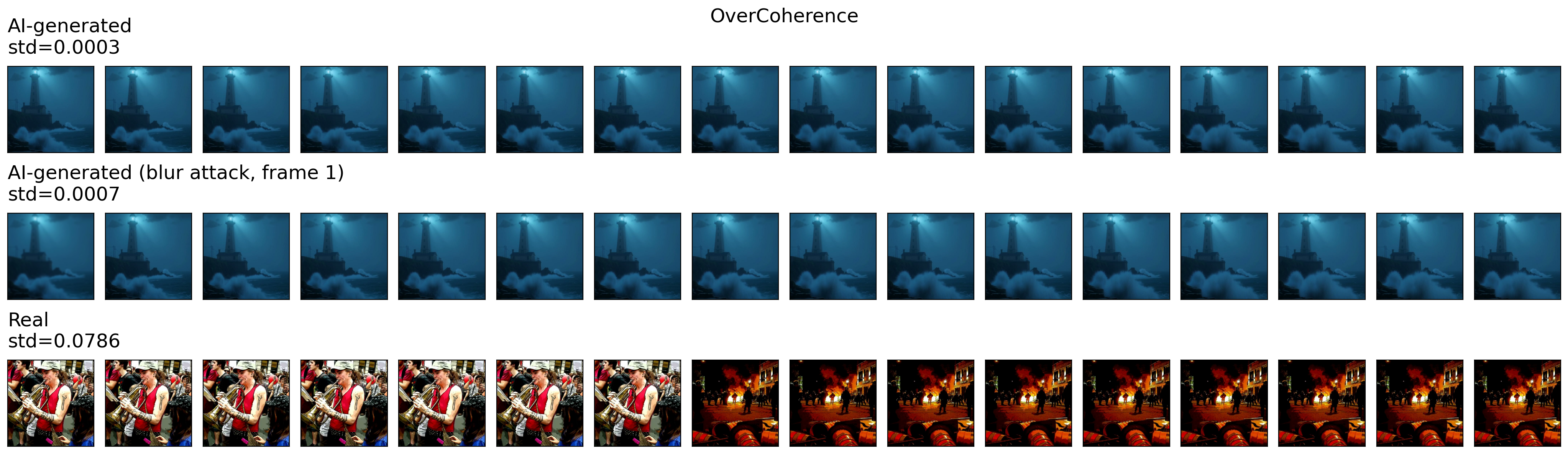}
    \includegraphics[width=\linewidth]{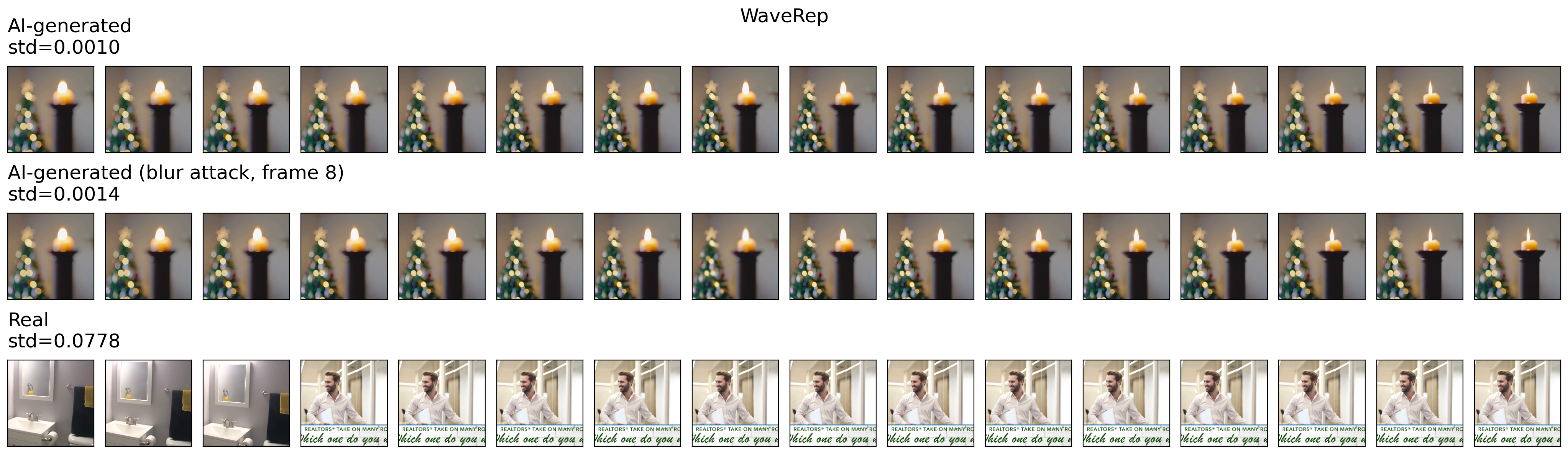}

    \caption{Examples of videos from the four biased datasets evaluated in Section \ref{sec:motion}, together with their motion std metric. AI-generated videos tend to have very limited motion, resulting in a low  standard deviation. In contrast, real videos generally contain more scene transitions and erratic motion, resulting in a high standard deviation. The middle rows show the result of the blur attack, where one randomly chosen frame is blurred. The visual effect this attack has on the video is minimal, while the resulting negative effect on the predictions of all evaluated models is substantial (Table \ref{tab:model_evaluation}).}
    \label{fig:appendix_motion_example}

\end{figure}

\begin{figure}[]
    \centering
    
    \includegraphics[width=\linewidth]{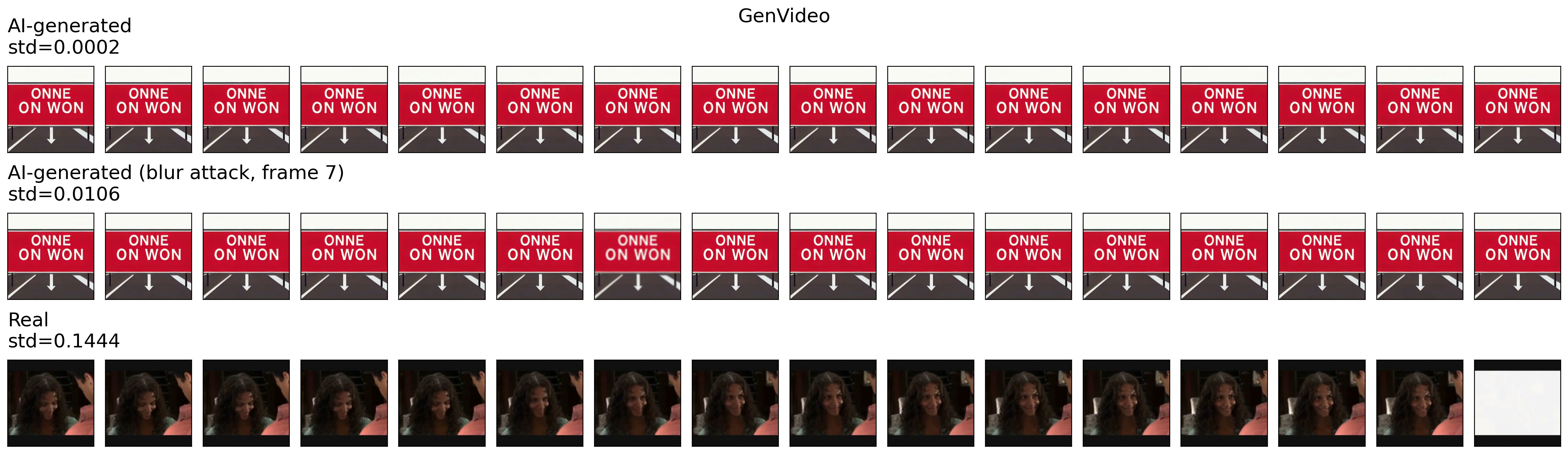}
    \includegraphics[width=\linewidth]{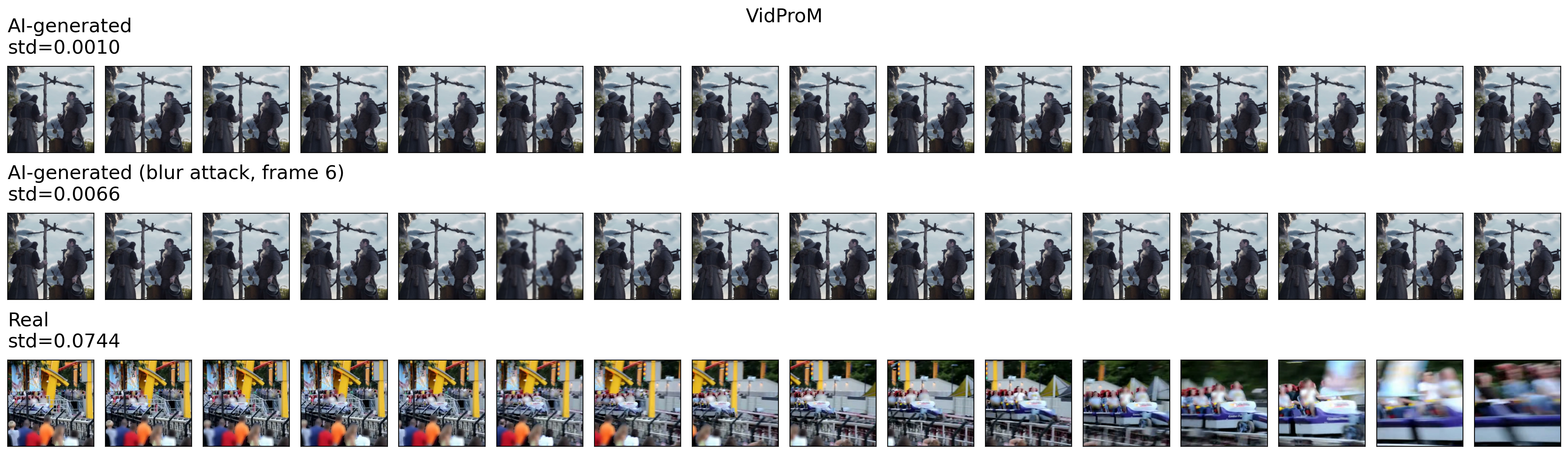}
    \includegraphics[width=\linewidth]{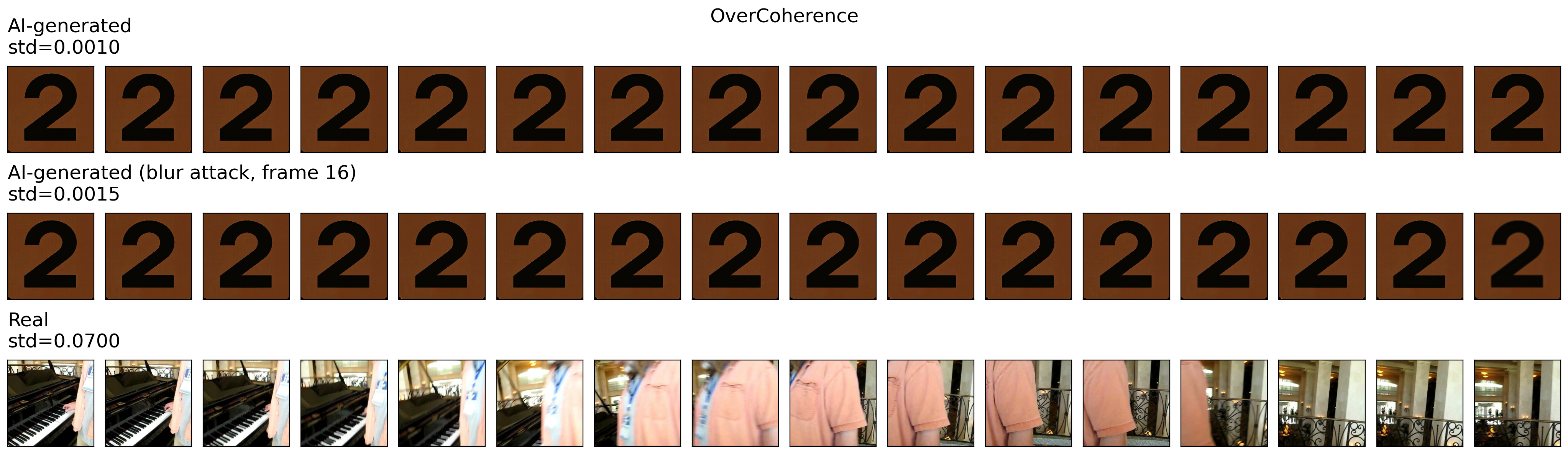}
    \includegraphics[width=\linewidth]{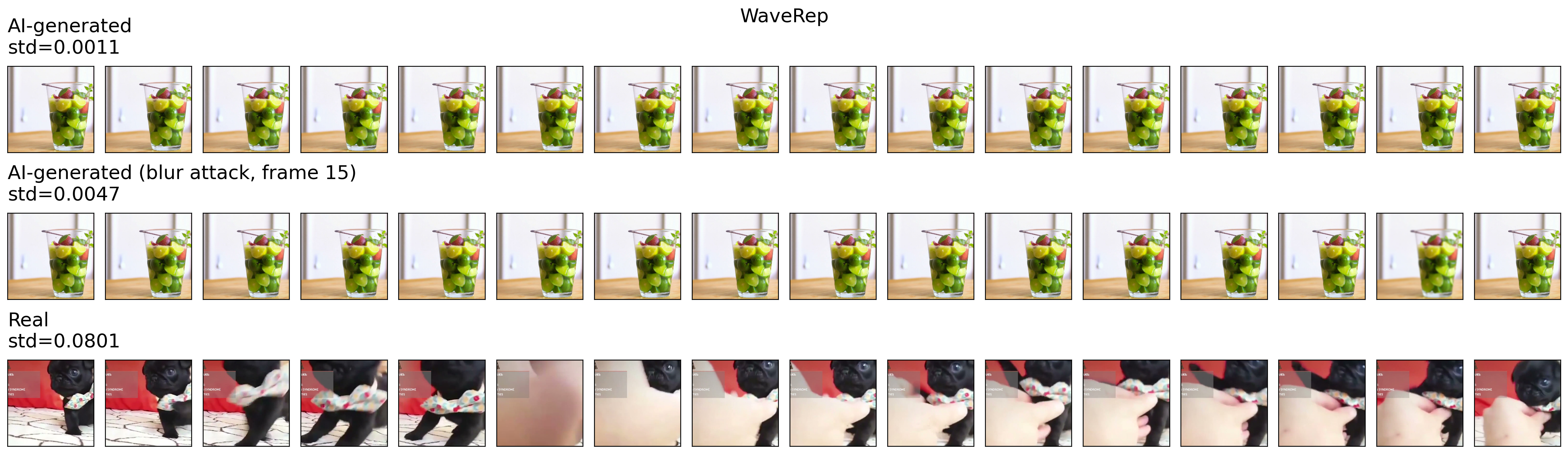}

    \caption{Additional examples of videos from the four biased datasets evaluated in Section \ref{sec:motion}, following the same structure as Figure \ref{fig:appendix_motion_example}.}
    \label{fig:appendix_motion_example_2}

\end{figure}

\clearpage

\subsection{Dataset licenses}
\def\UrlBreaks{\do\/\do\-\do\.\do\_\do\?\do\&\do\=}
\label{app:licenses}
\begin{itemize}
    \item \textbf{GenVideo}
    \begin{itemize}
        \item License: CC BY-NC 4.0
        \item \url{https://modelscope.cn/datasets/cccnju/Gen-Video}
    \end{itemize}
    
    \item \textbf{VidProM}
    \begin{itemize}
        \item License: CC BY-NC 4.0
        \item \url{https://huggingface.co/datasets/WenhaoWang/VidProM}
    \end{itemize}
    
    \item \textbf{Over-Coherence}
    \begin{itemize}
        \item License: Unknown
        \item \url{https://github.com/FujitsuResearch/training-free-detection-of-text-to-video-generations-via-over-coherence}
    \end{itemize}
    
    \item \textbf{WaveRep}
    \begin{itemize}
        \item License: Unknown
        \item \url{https://github.com/grip-unina/WaveRep-SyntheticVideoDetection}
    \end{itemize}
    
    \item \textbf{GenBuster++}
    \begin{itemize}
        \item License: MIT License
        \item \url{https://huggingface.co/datasets/l8cv/GenBuster-plusplus}
    \end{itemize}
\end{itemize}

\end{document}